\def\year{2021}\relax
\documentclass[letterpaper]{article} 
\usepackage{aaai21}  
\usepackage{times}  
\usepackage{helvet} 
\usepackage{courier}  
\usepackage[hyphens]{url}  
\usepackage{color,xcolor}
\usepackage{graphicx} 
\usepackage{natbib}  
\usepackage{amsmath,amssymb} 
\usepackage{caption} 
\usepackage{verbatim}

\title{Collaborative Distillation in the Parameter and Spectrum Domains \\
 for Video Action Recognition}

\author {
        Haisheng Su\textsuperscript{\rm 1},
        Jing Su\textsuperscript{\rm 1}\thanks{indicates equal contribution.},
        Dongliang Wang\textsuperscript{\rm 1},
        Weihao Gan\textsuperscript{\rm 1}, \\
        Wei Wu\textsuperscript{\rm 1},
        Mengmeng Wang\textsuperscript{\rm 2},
        Junjie Yan\textsuperscript{\rm 1},
        Yu Qiao\textsuperscript{\rm 3} \\
}
\affiliations {
    \textsuperscript{\rm 1} SenseTime Group Limited \\
    \textsuperscript{\rm 2} Zhejiang University \\
    \textsuperscript{\rm 3} Shenzhen Institutes of Advanced Technology, Chinese Academy of Sciences \\
    \{suhaisheng, sujing, wangdongliang, ganweihao, wuwei, yanjunjie\}@sensetime.com, \\
    mengmengwang@zju.edu.cn, yu.qiao@siat.ac.cn\\
}

\begin{document}

\maketitle

\begin{abstract}
Recent years have witnessed the significant progress of action recognition task with deep networks. However, most of current video networks require large memory and computational resources, which hinders their applications in practice. Existing knowledge distillation methods are limited to the image-level spatial domain, ignoring the temporal and frequency information which provide structural knowledge and are important for video analysis. This paper explores how to train small and efficient networks for action recognition. Specifically, we propose two distillation strategies in the frequency domain, namely the feature spectrum and parameter distribution distillations respectively. Our insight is that appealing performance of action recognition requires \textit{explicitly} modeling the temporal frequency spectrum of video features. Therefore, we introduce a spectrum loss that enforces the student network to mimic the temporal frequency spectrum from the teacher network, instead of \textit{implicitly} distilling features as many previous works. Second, the parameter frequency distribution is further adopted to guide the student network to learn the appearance modeling process from the teacher. Besides, a collaborative learning strategy is presented to optimize the training process from a probabilistic view. Extensive experiments are conducted on several action recognition benchmarks, such as Kinetics, Something-Something, and Jester, which consistently verify effectiveness of our approach, and demonstrate that our method can achieve higher performance than state-of-the-art methods with the same backbone.
\end{abstract}

\section{Introduction}
Nowadays, with rapid development of computer vision as well as increasing amount of digital cameras and Internet videos,  action recognition task has drawn much attention in the community and comes up with great progress \cite{feichtenhofer2016convolutional,wang2016temporal,sun2018optical,carreira2017quo,CPMN,liu2018t,tran2015learning}. And it aims to recognize the human actions in the manually trimmed videos.  However, convincing performance is usually achieved at the cost of vast parameters and resources, which limits their real-time applications on several areas, such as smart surveillance and video recommendation.

\begin{figure}[t]
	\centering
	\includegraphics[width=0.99\columnwidth]{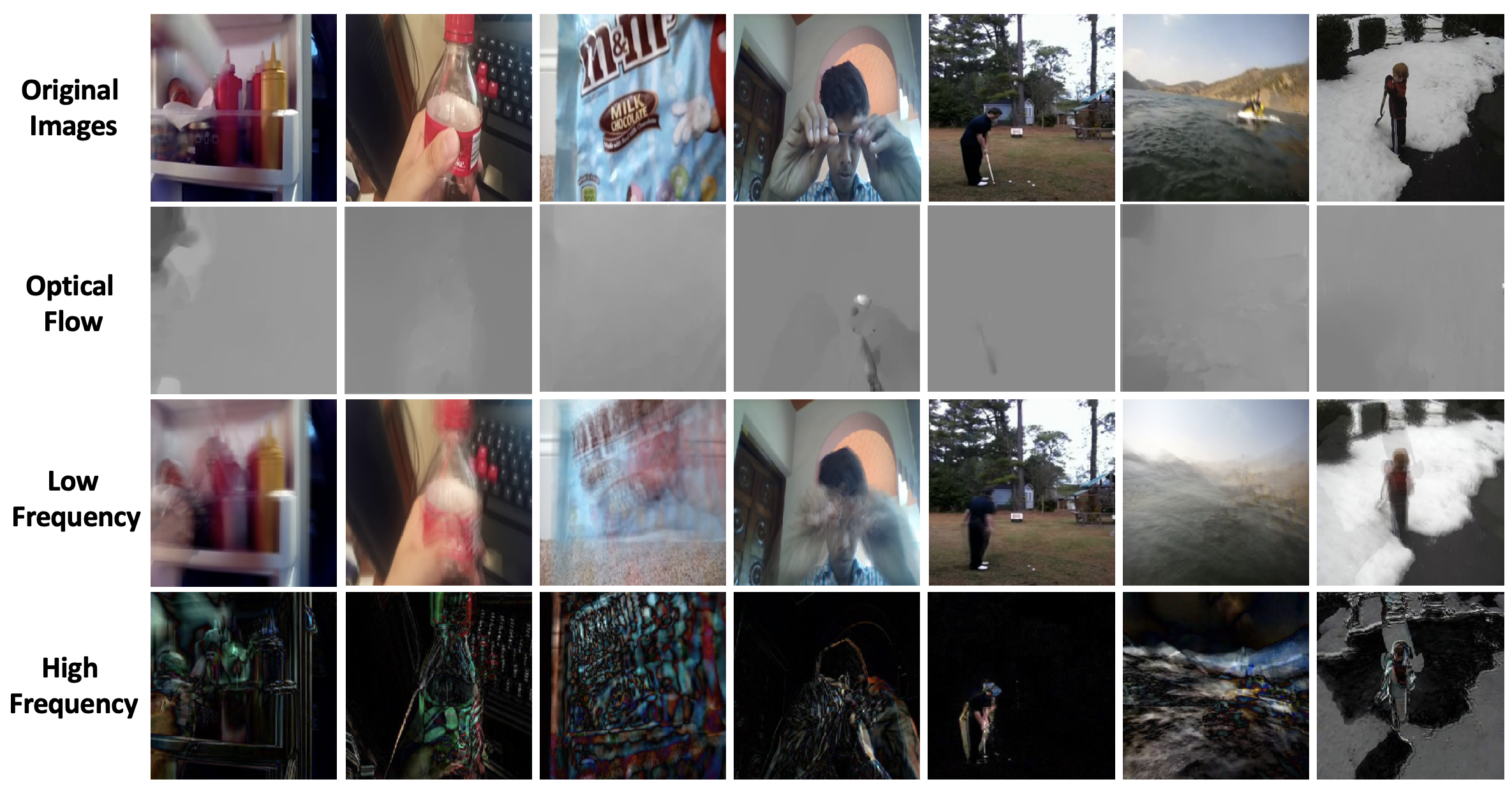}
	\caption{Illustration of the RGB images with different frequencies. The first row is the original images. The second row is the optical flow extracted by TV-L1.  The third row is the low frequency, which presents the scene representation. And the last raw is the high frequency, which describes the distinct motion edges.}
	\label{overview}
	\vspace{-0.4cm}
\end{figure}

Taking practicability into account, lightweight models with comparable performance attract many researchers in the community. To this aim, a sequence of advances \cite{feichtenhofer2016convolutional,tran2015learning,wang2018appearance} elaborately design convolutional networks to integrate both spatiotemporal and motion features into a unified 2D CNN framework, without any 3D convolution and optical flow pre-calculation which will inevitably increase the computing cost. However, these action classifiers still rely on large and deep backbone to achieve promising results. Knowledge distillation has been studied widely and shown feasibility in the image-level task \cite{thoker2019cross,hinton2015distilling,heo2019knowledge,park2019relational}, which aims to train small and efficient video networks (student) with comparable performance as large ones (teacher). However, different from image-level distillation, for action recognition in videos, there are two crucial and complementary cues: \textbf{appearances} and \textbf{temporal dynamics}. Bhardwaj et al. \cite{bhardwaj2019efficient} and Farhadi et al. \cite{farhadi2019tkd} aim to perform data distillation for key frames selection as video input, without considering the data-invariant network distribution. Girdhar et al. \cite{girdhar2019distinit} directly borrow the idea of image-level distillation, neglecting the temporal inter-dependencies between neighboring frames.


To address the above issues, we propose two distillation strategies in the frequency domain for video action recognition, namely the \textit{feature spectrum distillation} and \textit{parameter distribution distillation}. As shown in Fig. \ref{overview}, we illustrate the original RGB frames and corresponding different frequencies. We observe that the low frequency usually pays attention to the scene appearance while the high frequency describes the motion information with the distinct edges similar to the optical flow. In order to capture both temporal dynamics and scene appearance through different frequencies, we calculate the feature frequency spectrum along the temporal domain, and then adopt a spectrum loss to perform mimic learning from teacher networks. Moreover, since the convolution operation in the spatial domain has been proven to be equivalent to the multiplication in the frequency domain \cite{bracewell1986fourier}, and the distribution of network parameter can provide data-invariant structural knowledge used for feature extraction. Under this observation, we further propose the parameter distillation method which adopts the parameter frequency distribution to guide the student network to learn spatial modeling process from the teacher network. Finally, we exploit the collaborative learning strategy during the distillation process to eliminate the redundant and dark knowledge dynamically from the teacher model.
 
In summary, our main contributions are three-folds:

(1) To the best of our knowledge, we are the first to perform video knowledge distillation in the frequency domain for action recognition, where temporal feature spectrum and network parameter distribution are both considered for training small and efficient models.

(2) We propose two distillation strategies to guide the learning of temporal dynamics and appearance modeling process from teacher networks respectively. Besides, the collaborating learning strategy is adopted to eliminate the redundant knowledge dynamically for efficient distillation.

(3) Extensive experiments are conducted on several action recognition benchmarks, such as Kinetics, Something-Something, and Jester, which confirms the effectiveness and efficiency of our knowledge distillation method.



\section{Related Work}

\noindent
{\bf Action Recognition.}  Action recognition is an important branch of video analysis area which has been widely explored in recent years. Earlier methods manually design hand-crafted features, while recent deep learning based methods learn the features automatically. Current literature can be mainly divided into two main categories. A sequence of advances \cite{feichtenhofer2016convolutional,wang2016temporal} adopt two-stream networks to capture the appearance features and motion information from RGB images and stacked optical flow respectively. 3D networks  \cite{tran2015learning,carreira2017quo} exploit the 3D convolution to capture spatial and temporal information directly from raw videos. However, 3D convolution inevitably increases computing cost which limits its real-time applications. Under this circumstance, several methods seek for the trade-off between the accuracy and speed. Tran et al. \cite{tran2019video} and Qiu et al. \cite{P3d} propose to decompose the 3D convolution into 2D spatial convolution and 1D temporal convolution. Lin et al. \cite{lin2018temporal} shift part of the channels along the temporal dimension to enable the temporal information transmission between neighboring frames. Jiang el al. \cite{jiang2019stm} further integrate both spatiotemporal and motion features into a unified 2D CNN framework, without any 3D convolution and optical flow pre-calculation. 


\begin{figure*}[t]
        \centering
	\includegraphics[width=14cm]{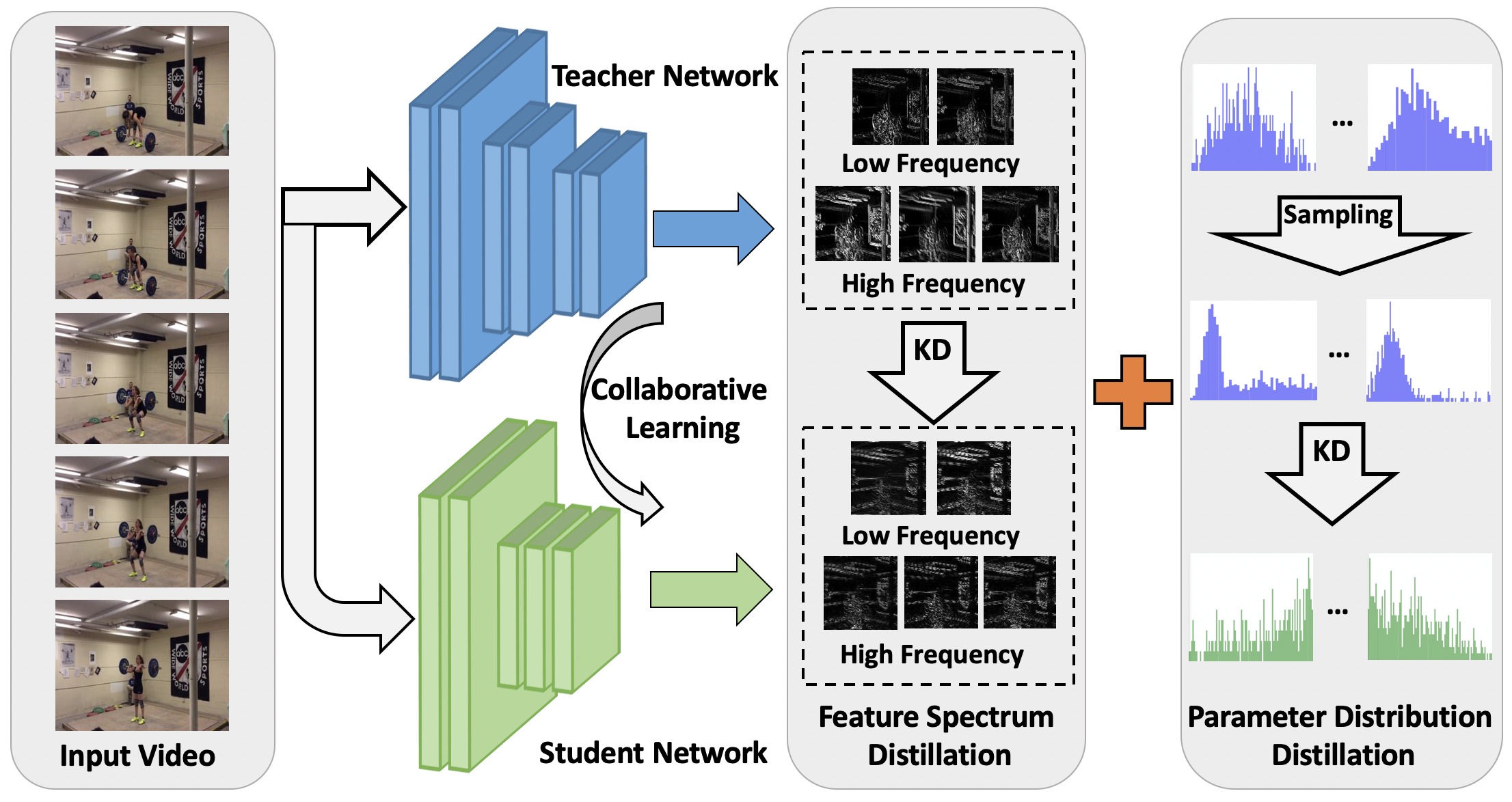}
	\caption{The framework of FPCD. To mimic the motion and scene representation from teacher network, the frequency spectrum is computed based on the video features between neighboring frames. Besides, the sampled frequency distribution of network parameters are also distilled for appearance modeling, which is independent of input video data. Finally, the two distillation losses ( i.e. spectrum loss and $KL$-divergence loss) are combined with the standard classification loss through the collaborative learning strategy, thus eliminating the redundant knowledge adaptively and dynamically from the teacher network.}
	\label{framework}
\end{figure*}

\noindent
{\bf Knowledge Distillation.} An important branch of this field is to leverage the useful knowledge from deep network to train a lightweight model. \cite{ZagoruykoK16a,Yim2017A} perform knowledge distillation for image classification by using the class probabilities as a soft target or directly distilling the features into student model. Recently, the vast majority of methods manage to effectively learn the intermediate features, including both local and global information \cite{park2019relational}.  \cite{Rongcheng,Wang_2019_CVPR,AkopyanLarge} explore the relations between the feature maps on both sides, which are actually abstract and hard to define. A graph is generated by the Multi-Head Attention (MHA) \cite{Seunghyun} which provides a relational inductive bias to improve the performance significantly. To further distill diverse knowledge, the proposed ensemble method (MEAL) \cite{shen2019meal} adopts an adversarial-based learning strategy. Meanwhile, the dubbed Relational Knowledge Distillation (RKD) \cite{park2019relational} transfers the distance-wise and angle-wise relations from the teacher to student networks. 

For other image-level tasks, there are also several works showing encouraging results. One of the major challenges of lane detection task is to train deep models. Self-Attention Distillation (SAD) \cite{hou2019learning} designs a model to learn from itself and thus gains improvement. As for the semantic segmentation task, the dilemmas of heavy computational cost and inferior efficiency remain to be solved. Most of methods directly perform knowledge distillation on each pixel separately, neglecting the important structure information. He et al. \cite{he2019knowledge} handle the inconsistent features through modeling the feature similarity in a transferred latent domain to capture the long-range dependencies. 

The existing video analysis algorithms usually require for large-scale data transmission, several works focus on how to use relative fewer input frames for action recognition while maintaining competitive accuracy \cite{LuoGraph,Garcia2018Modality}. Memory-efficient cluster-and-aggregate models only look at a small fraction of frames in the video in contrast to a typical Teacher-Student setting. Temporal Knowledge Distillation (TKD) \cite{farhadi2019tkd} framework selects video frames of the perception moments and adopts a light-weight model to obtain a convincing performance.  However, this kind of LSTM-based key frame selection methods would hinder the real-time application. Temporal Sequence Distillation (TSD) \cite{huang2018knowledge} also employs a light-weight model to distill a long video sequence into a very short one for action recognition.

\noindent
{\bf Frequency Domain Learning.}  Most of frequency-based methods aim to reduce the computing cost and parameters with Fourier Transformation (FT), thus improving the network efficiency. Winograd-based and FFT-based approaches improve the running time and the processor’s speed at the same time \cite{jia2018optimizing}.  An efficient design of FFT-based neural architecture reveals that through merely replacing the point-wise convolutions, significant performance can be achieved. Wang et al. \cite{wang2018superneurons} focus on the sparse significant of gradients in the frequency domain to further compress gradients frequencies, and the proposed compression framework effectively improves the scalability of most popular neural networks. Besides, the proposed method also emphasizes that the networks designed in the frequency domain are usually of high efficiency. In this paper, we elaborately design a light-weight model with \textit{feature spectrum} and \textit{network parameter} distillations in the frequency domain, to mimic both temporal dynamics and appearance clues from teacher networks in a collaborative learning way. Extensive experiments demonstrate that our method can achieve compelling performance with different video backbones on several public benchmarks respectively.


\section{Our Approach}
In this section, we will introduce the technical details of our proposed method. The framework is shown in Fig. \ref{framework}. To begin with, we denote an input video as  $\mathbf{V}$ with size of $\mathbb{R}^{ T \times W \times  H \times 3 } $, where $W$ and $H$ are the width and height of an input image respectively, and $T$ indicates the number of frames. Then we extract the output features $\mathbf{F}$ of size $\mathbb{R}^{ T \times W^{'} \times H^{'} \times C} $ from different stages separately and then compute their frequency spectrums along the temporal domain as $\ \hat{\mathbf{F}}$ of size $\mathbb{R}^{ K \times W^{'} \times H^{'} \times C} $, where $C$ is the channel number and $K$ is the number of frequency bands. 

To achieve high-quality distillation for action recognition, we propose the Feature and Parameter Collaborative Distillation (FPCD) method with two distillation strategies, namely the Feature Spectrum Distillation (FSD) and Parameter Distribution Distillation (PDD). \textit{Feature spectrum distillation} is performed upon the output features in the frequency domain among different stages of the network backbone, which captures the motion structure and scene representations along the temporal dimension for mimic learning. \textit{Parameter distribution distillation} further aligns the parameter frequency distribution between the teacher and student networks, thus guiding the appearance modeling process of the student network, which is irrelevant of the video datasets. Finally, the two distillation terms and the standard classification loss are jointly optimized in a collaborative learning manner, where redundant and negative knowledge are adaptively eliminated during the distilling process.


\subsection{Feature Spectrum Distillation} 

Existing video action classifiers (i.e., TSN\cite{wang2016temporal}, STM\cite{jiang2019stm}, I3D\cite{carreira2017quo}, etc.) learn unstructural features of scene and motion information, which are not conducive to transfer from teacher to student networks for mimic learning. As shown in Fig. \ref{overview}, we find that the frequency spectrum computed based on the input video frames can present the motion and scene information with different frequencies. Specifically, high frequency attends to the motion information between neighboring frames, while the low frequency pays attention to the scene representation. Therefore, frequency spectrum can be served as a kind of structural knowledge for distillation between teacher and student networks, which is easier for the student training process with faster converge speed. 

\begin{figure}[t]
        \centering
        \includegraphics[width=8cm]{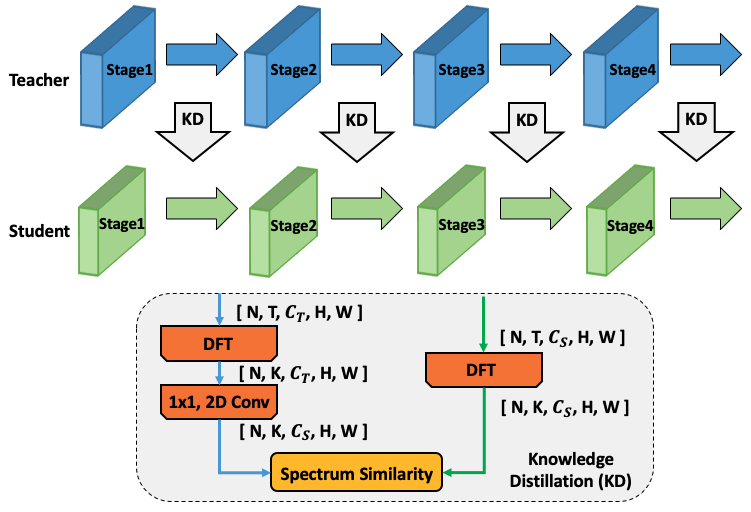}
	\caption{Detailed structure of Feature Spectrum Distillation (FSD) module. With Discrete Fourier Transform (DFT), the temporal frequency spectrums are computed based on the output features from different stages separately. And a 2$D$ convolution is adopted as a predictor for channel reduction in order to ensure the consistent output shape on both sides for spectrum similarity calculation.}
	\label{FSD}
	\vspace{-0.3cm}
\end{figure}

The overview of the feature spectrum distillation process is shown in Fig. \ref{FSD}. Given a video $V$, we use ResNet $Net_s$ to extract feature $\textbf{F}_{s}^{i}$, and then we apply the Discrete Fourier Transform (DFT) to transfer the extracted features of student network from temporal domain to frequency domain:
\begin{equation}
\begin{split}
    & \textbf{F}_{s}^{i}= Net_s(\textbf{V},\theta_s) , \\
    & \hat{\textbf{F}}_{s}^{i}= DFT(\textbf{F}_{s}^{i}) ,
\end{split}
\end{equation}

\noindent
where $\hat{\textbf{F}}_{s}^{i}$ is the computed frequency spectrum. 
$ \theta_s $ denotes the learnable parameters. Similarly, we compute the frequency spectrums of the teacher network based on the output feature maps in different stages, which are then employed as distillation knowledge. Unlike the previous methods which perform distillation by sampling intermediate features from the teacher model, we adopt a predictor which contains a 2$D$ convolution with kernel size of 1 for channel reduction from $C_{T}$ to $C_{S}$ as shown in Fig. \ref{FSD}, thus ensuring the consistent shape of the output frequency spectrums on both sides for distillation. In this way, feature spectrum extracted from the student network in the frequency domain is fed to the predictor for distilling from the teacher network with the spectrum loss $L_S$: 
\begin{equation}
    L_S(\hat{\textbf{F}}_{t}^{i}, \hat{\textbf{F}}_{s}^{i})=-\frac{1}{\left | \chi \right |}
    \sum\limits_{(\hat{\textbf{F}}_{t}^{i},\hat{\textbf{F}}_{s}^{i})\in \chi} (\hat{\textbf{F}}_{t}^{i}) \log (f(\hat{\textbf{F}}_{s}^{i}, \theta_p)) , 
\end{equation}
where $\hat{\textbf{F}}_{t}^{i}$ and $\hat{\textbf{F}}_{s}^{i}$ are the teacher and student frequency spectrums in the $i^{th}$ stage of the ResNet backbone respectively, and $\chi$ is the mini-batch data of size $\left | \chi \right |$. $f(\cdot)$ is the predictor model with learnable parameters $ \theta_p $. During the training phase, the predictor together with the student network is optimized simultaneously. 


Mathematically, we derive the frequency spectrum based on the video features to explicitly explain the relationship between these two domains. The expansion of DFT function is written as:
\begin{equation}
\begin{split}
& \hat{\textbf{F}}[k] = \sum\limits_{p=0}^{T-1} {\textbf{F}}(p) e^{-j \frac{2 \pi}{T} kp} , (k=0, 1, 2, \cdots T-1)
\end{split}
\end{equation}

For the convenience of discription, here we set $T=2$, where $k=1$ denotes the high frequency, and $k=0$ denotes the low frequency:
\begin{equation}
 \begin{split}
& \hat{\textbf{F}}[0] = \textbf{F}(0) + \textbf{F}(1) ,  \\
&\hat{\textbf{F}}[1] = \textbf{F}(0) - \textbf{F}(1) .
  \end{split}
\end{equation}

We can conclude that the low-frequency representation $\hat{\textbf{F}}[0]$ is the sum of video features (i.e., $\textbf{F}(0)$ and \textbf{F}(1)), while the high-frequency representation is the difference between neighboring video features. Therefore, the low-frequency representation can retain the most of scene information. While in the high frequency, the scene information would be counteracted, and the distinct motion edges would be highlighted. The visualization result is illustrated in Fig. 3.

\subsection{Parameter Distribution Distillation}

Most feature-based knowledge distillation methods rely on the training data greatly, ignoring the characteristics of the neural network itself. However, Bracewell et al. \cite{bracewell1986fourier} claim that the convolution operation in the spatial domain approximates the multiplication of spatial feature frequency and 2$D$ CNN parameter frequency:
\begin{equation}
    \begin{split}
     CNN(\textbf{V}, \theta_s) \cong DFT(\textbf{V}) \times  DFT(\theta_s), 
    \end{split}
\end{equation}
where $CNN(\cdot)$ and $DFT(\cdot)$ indicate the 2$D$ convolutional network and Discrete Fourier Transform (DFT) function respectively. And $\textbf{V}, \theta_s$ represent the input video and network parameter respectively. Therefore, the convolution searches over different parameter frequencies to extract specific frequency of spatial features. Motivated by this observation, we propose the parameter distribution distillation strategy in the frequency domain.

Specifically, the network parameters are optimized to extract key and discriminative video features for action recognition. Meanwhile, the convolution process aims to capture different features of frequency. Therefore, the parameter weights serve as a selector among different input frequencies. It is intuitive and important to adopt the parameter frequency distribution as distillation knowledge for teaching the student networks. As shown in Fig. \ref{framework}, we  first downsample the parameters of the teacher network to the same number of student ones, then we sort the high-frequency and low-frequency parameters on both sides, and adopt the $KL$-divergence loss $L_{P}$ for knowledge distillation between the teacher and student networks:
\begin{equation}
    \begin{split}
        L_{P}(\theta_t,\theta_s) &= \frac{1}{|\mathcal M|} 
\sum\limits_{(\theta_t,\theta_s)\in \mathcal M} KL(PD_s || PD_t) ,
    \end{split}
\end{equation}
where $PD_{s}$ and $PD_{t}$ indicate the parameter distribution of student and teacher networks respectively, where $PD_{s} = DFT(\theta_s)$, $PD_{t} = DFT(g(\theta_t))$. And g($\cdot$) denotes the sampling function, which randomly downsamples the same number of convolutional kernels during distillation process. $\theta_s$, $\theta_t$ denote the parameters of student and teacher networks respectively. $|\mathcal M|$ is the number of convolutional kernels $\mathcal M$.


\subsection{Collaborative Learning}

During the distillation process, there will inevitably exist misinformation in the teacher network, namely \textit{dark knowledge}. Undoubtedly, these knowledge doesn't make sense and should not be distilled into the student network, which may mislead the student for action recognition task. Thus the collaborative learning strategy is introduced for effective distillation from a probabilistic view, in order to select the valid information adaptively from the teacher network based on the output confidence.

\begin{figure}[t]
	\setlength{\abovecaptionskip}{-0.2cm} 
	\begin{center}
		\includegraphics[width=0.8\linewidth]{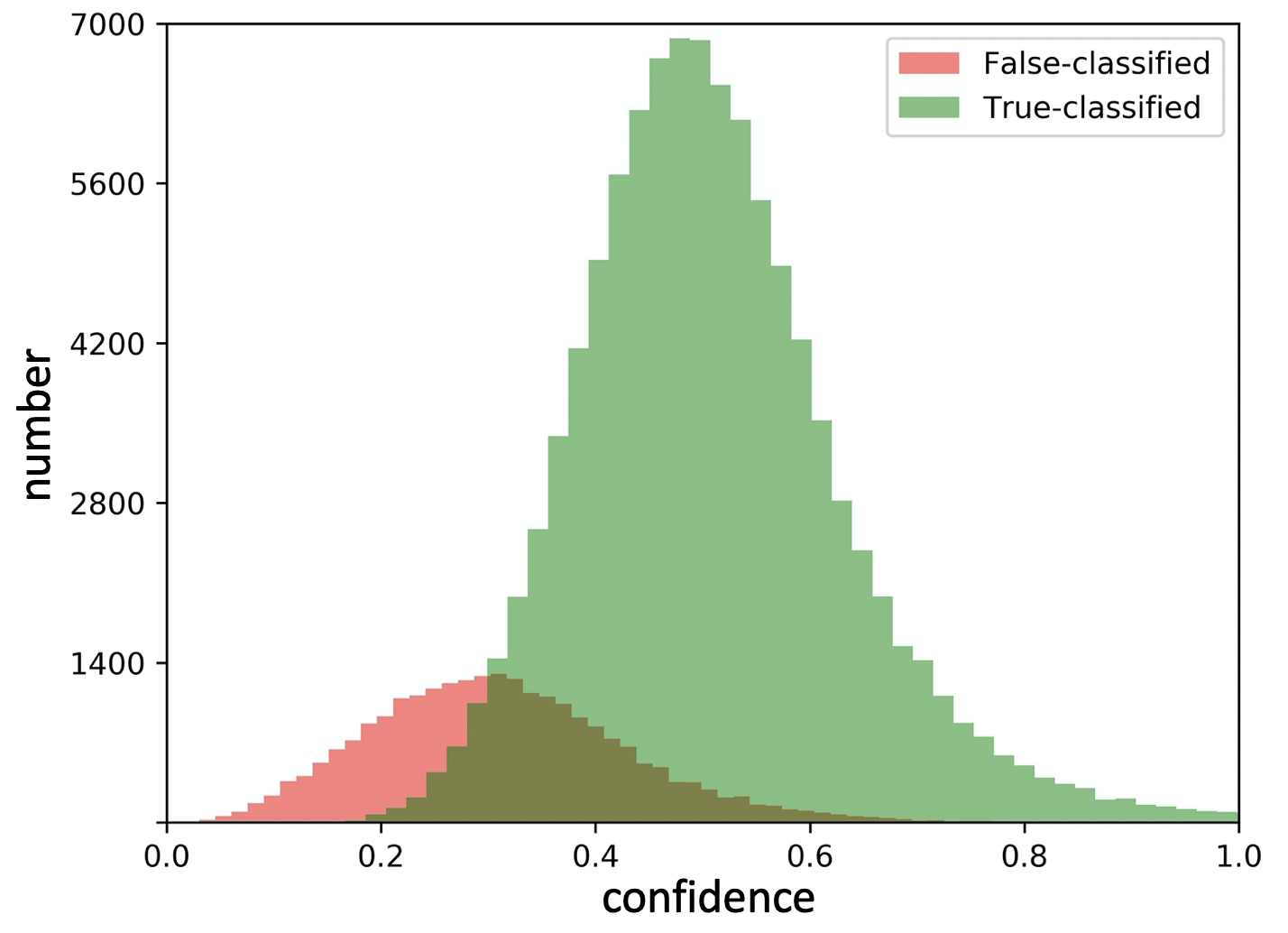}
	\end{center}
	\caption{Illustration of output confidence distributions of the teacher network for true-classified and false-classified samples respectively on Something-Something V2 dataset.}
	\label{cl_distribution}
	\vspace{-0.2cm}
\end{figure}

Specifically, as illustrated in Fig. \ref{cl_distribution}, we visualize the true-classified output confidence distribution and the false-classified output confidence distribution of the teacher network on a specific dataset (i.e. Something-Something V2) respectively. We denote the output confidence of the teacher network as $c_{t}$, and $P(c_{t})$ is the proportion of the output confidence $c_{t}$ relative to the overlapped area. The probability $P'(c_{t})$ denotes the selecting probability with output confidence $c_{t}$:
\begin{equation}
    \begin{split}
    &P(P|T)    = P(c_{t}) * P'(c_{t}) ,  \\
    &P(P|F)   =  (1 - P(c_{t})) *  P'(c_{t}) , \\
    &P(N|T)   = P(c_{t}) * (1 - P'(c_{t})) ,  \\
    &P(N|F)  = (1 - P(c_{t})) * (1 - P'(c_{t})) , 
    \end{split}
\end{equation}
where $P(P|T)$ denotes the probability of selecting features of a true-classified sample, and $P(P|F)$ denotes the probability of selecting features of a false-classified sample. Similarly,  $P(N|T)$ denotes the probability of ignoring the true-classified video features, while $P(N|F)$ denotes the probability of ignoring the false-classified video features. Obviously, our goal is to maximize the value of  $P(P|T) + P(N|F)$ as:
\begin{equation}
    \begin{split}
    & P(P|T) + P(N|F) \\
    &= P(c_{t}) P'(c_{t}) + (1 - P(c_{t})) * (1 - P'(c_{t})) \\
    &= (2P(c_{t}) - 1)  P'(c_{t}) + (1 - P(c_{t}))
    \end{split}
\end{equation} 
where $(1 - P(c_{t}))$ is a constant determined by the output distribution of a specific dataset, hence the value of $P'(c_{t})$ is determined by the value of $(2P(c_{t}) - 1)$, then we can obtain:
\begin{equation}
    P'(c_{t}) = \left\{\begin{matrix}
    1 \quad  & c_t > \theta_{P(c_{t}) = 0.5} \\ 
    0  \quad & c_t \leq \theta_{P(c_{t}) = 0.5} \\ 
\end{matrix}\right.     
\end{equation} 

We divide the knowledge distillation process into three stages against training epochs. Concretely, the weight factor in the first $N_1$ epochs could be defined as a constant with the probability $P'(c_{t})$. Then in the following $N_2 - N_1$ epochs, the weight factor could be dynamically changed with an exponential function. No matter what stage, we maintain the sampling principle based on the output confidence distribution of teacher network. To implement the collaborative learning strategy, we start with setting the weight of high value to select teacher features and network parameters for distillation, then the weight is decreased by the exponential decay. And in the last stage, the weight factor would be a relatively small constant. The weight factor changes with training epochs which can be described as follows:
\begin{equation}
   f(n)=\left\{\begin{matrix}
    \gamma \cdot P'   \quad & n \leq N_{1} , \\ 
    (\lambda^{n-N_{1}} + \alpha) \cdot P'  \quad & N_{1}<n\leq N{_{2}} , \\ 
    \alpha \cdot P'   \quad & n>N{_{2}} ,
\end{matrix}\right. 
\end{equation}
where $\gamma$, $\alpha$, and $\lambda$ denote constant weights respectively. $n$ denotes the training epoch, $N{_{1}}$ denotes the number of epochs in the first stage, and  $N{_{2}}$ denotes the maximum number of epochs in the second stage. Empirically, $\gamma$, $\alpha$, and $\lambda$ are set to 0.9, 0.1 and 0.8 separately.

\subsection{Overall Loss Function}
As described above, FPCD consists of three main optimization terms, namely the standard classification loss $L_{Cls}$, the spectrum loss $L_{S}$ and the parameter $KL$ loss $L_{P}$. The multi-task objective function is defined as:
\begin{equation}
L_{FPCD} = L_{Cls} + f(n) \cdot (L_{P} + L_{S}), 
\end{equation}
where $f(n)$ indicates the balance weight factor defined in the former section. $L_{Cls}$ is a cross entropy loss.

\section{Experiments}


{\subsection{Dataset and setup} 
{\bf Dataset.} We evaluate the performance of proposed FPCD on several public action recognition datasets, including temporal-related datasets (i.e. Something-Something V1 \& V2 \cite{goyal2017something} , Jester) and scene-related dataset (i.e. Kinetics-400 \cite{carreira2017quo}). Kinetics-400 is a large-scale human action video dataset with 400 classes which contains 236,763 clips for training and 19,095 clips for validation. Something-Something V2 contains 174 classes with 220,847 videos which is larger than Something-Something V1 with 108,499 videos. Jester is a relatively small dataset with only 27 classes and 148,092 videos for generic human hand gesture recognition. For the temporal-related datasets, the temporal motion information and the interaction with objects are important for action recognition. Most of the actions cannot be correctly recognized without considering these clues. While for the scene-related datasets, the background information contributes the most for determining the action category, hence the temporal information is not as important as the former one.

\noindent
{\bf Implementation details.}  We perform the distillation regularization terms on the each stage output of network backbone (i.e. ResNet) between the teacher and student networks. Meanwhile, we set different weights for each stage (i.e., 1/8, 1/4, 1/2, 1) respectively. Instead of adopting the knowledge distillation for all stages, we randomly select one stage during the training phase. We use ResNet-18 \cite{K.He} as the backbone. The sparse sampling strategy \cite{wang2016temporal} is utilized to extract $T$ frames from the video clips ($T$  = 8 or 16 in our experiments, 8 by default without specific explanation). During training, random scaling and corner cropping are utilized for data augmentation, and the cropped region is resized to 224 $\times$ 224 for each frame.

During the test phase, the efficient protocol (center crop $\times$ 1 clip) is considered for method evaluation, in which 1 clip with $T$ frames is sampled from the video. Each frame is resized to 320 $\times$ 256, and a central region of size 224 $\times$ 224 is cropped for action prediction.


\subsection{Comparison with the state-of-the-art methods}
Comparison with other methods of the similar backbone depth (i.e. ResNet-18) on Kinetics-400 dataset is shown in Table \ref{sota}. We can observe that: (1) our FPCD can effectively improve the performance of existing action recognition frameworks (i.e. TSN \cite{wang2016temporal} and STM \cite{jiang2019stm}) with different number of input frames ($T$  = 8 or 16). (2) With our distillation method, our action classifier can achieve the state-of-the-art performance compared with other methods, including 3D-ResNet \cite{hara2018can} , C3D \cite{tran2015learning} and ARTNet \cite{wang2018appearance}, which demonstrate the consistent effectiveness and generalizability of our knowledge distillation method for training a lightweight video network. 


\setlength{\tabcolsep}{4pt}
\begin{table}[t]
    \setlength{\belowcaptionskip}{-0.2cm} 
    \setlength{\abovecaptionskip}{-0.1cm} 
    \caption{Comparison of different action recognition methods with similar backbone size (i.e. ResNet-18) on Kinetics-400.}
        \begin{center}
        \begin{tabular}{ccccccc}
        \hline
        Method & Backbone & Input & Top-1 & Top-5 & Avg \\
        \hline\hline
        3D-ResNet  &  ResNet-18 & 16 &   54.2  &   78.1 &  66.1   \\
        C2D  & ResNet-18 & 16 & 61.2 & 82.6 &   71.9     \\
        C3D  &  - & 16 & 65.6 &  85.7 & 75.7    \\
        TrajectoryNet &ResNet-18  & 16 & - & - & 77.8 \\
        ARTNet  &ResNet-18   &  16 & 67.7 & 87.1 & 77.4 \\
        TSN  &ResNet-18   &  8 & 61.4 &  83.4  & 72.4   \\
        TSN  &ResNet-18   &  16 &  62.5 &  84.0  & 73.3   \\
        STM  &ResNet-18   &  8 & 64.4 &  85.8  & 75.1   \\
        STM  &ResNet-18   &  16 & 66.9 &  87.6  & 77.2  \\
        \hline
        \textbf{FPCD} + TSN & ResNet-18 & 8 & 65.3 &  86.7  & 76.0 \\
        \textbf{FPCD} + TSN & ResNet-18 & 16 & 66.7 &  87.1  & 76.9 \\
        \textbf{FPCD} + STM & ResNet-18 & 8 & 69.4 &  89.1 &  79.4 \\
        \textbf{FPCD}  + STM& ResNet-18 & 16 & \textbf{70.3} &  \textbf{89.9} & \textbf{80.1} \\
        \hline       
    \end{tabular}
    \end{center}
    \label{sota}
    \vspace{-0.1cm}
\end{table}

\setlength{\tabcolsep}{1pt}
\begin{table}[t]
    \setlength{\belowcaptionskip}{-0.1cm} 
    \setlength{\abovecaptionskip}{-0.1cm} 
    \caption{Top-1 accuracy comparison between FPCD and other distillation methods (i.e. Simple KD \cite{hinton2015distilling} and CCKD \cite{2019Correlation}) on Something-Something, Jester and Kinetics-400 datasets separately with STM \cite{jiang2019stm} framework.}
        \begin{center}
        \begin{tabular}{ccccccc}
        \hline
         Method &Backbone & SS-V1& SS-V2 & Jester & Kinetics \\
        \hline\hline
        STM & ResNet-50  &    47.5 &     60.4      &  96.6 & 71.1    \\
        \hline
        STM & ResNet-18  &  39.0   &  55.2     &  95.1 &  64.4    \\
        Simple KD + STM & ResNet-18  &    44.6        &       57.6       &  95.6 & 67.1   \\
        CCKD + STM & ResNet-18  &    45.1       &     58.7       &  96.0 & 67.7  \\
        \textbf{FPCD} + STM & ResNet-18 & \textbf{46.7} &\textbf{60.6} & \textbf{96.1} & \textbf{69.4} \\
          \hline       
    \end{tabular}
    \end{center}
    \label{stm}
   \vspace{-0.3cm}
\end{table}

\setlength{\tabcolsep}{4pt}
\begin{table}[t]
 \setlength{\belowcaptionskip}{0.1cm} 
  \setlength{\abovecaptionskip}{-0.1cm} 
      \caption{Comparison between low-frequency and high-frequency distillations of Feature Spectrum Distillation (FSD) with STM \cite{jiang2019stm} on Kinetics-400 dataset.}
    \begin{center}
    \begin{tabular}{p{4.2cm}ccc}
        \hline
        Method & Backbone & Top-1 & Top-5 \\ 
        \hline\hline
        FPCD (w/o FSD-low) + STM & ResNet-18   &     67.7       &     88.6    \\
        FPCD (w/o FSD-high) +STM & ResNet-18    &   69.0       &  88.9    \\
        \hline
        \textbf{FPCD} + STM & ResNet-18    &     \textbf{69.4}       &     \textbf{89.1}   \\
        \hline       
    \end{tabular}
    \end{center}
    \label{fsd}
   \vspace{-0.4cm}
\end{table}

\setlength{\tabcolsep}{3pt}
\begin{table}[t]
   \setlength{\belowcaptionskip}{0.3cm} 
    \setlength{\abovecaptionskip}{-0.1cm} 
    \caption{Top-1 accuracy comparison of different modules proposed in FPCD with STM \cite{jiang2019stm} framework on different datasets. FSD denotes the feature spectrum distillation, PDD denotes the parameter distribution distillation, and CL denotes the collaborative learning.}
    \begin{center}
    \begin{tabular}{p{2.8cm}cccc}
        \hline
        Method & Backbone & SS-v1   & SS-v2 & Kin-400 \\
        \hline        
        \hline
        Original STM     & ResNet-18 &       39.0       &    55.2   &      64.4              \\
         \hline
         + FSD    & ResNet-18 &          44.8       &           59.0         &       65.8              \\
        + PDD     & ResNet-18 &       43.2       &           58.7         &        66.6             \\
        + FSD + PDD   & ResNet-18  &      45.6        &           59.3          &      67.5               \\
        + FDD + PDD + CL    & ResNet-18 & \textbf{46.7}  &  \textbf{60.6}    & \textbf{69.4}      \\
        \hline       
    \end{tabular}
    \end{center}
    \label{fpcd}
   \vspace{-0.6cm}
\end{table}

\begin{figure}[t]
\centering
    \includegraphics[width=0.45\textwidth]{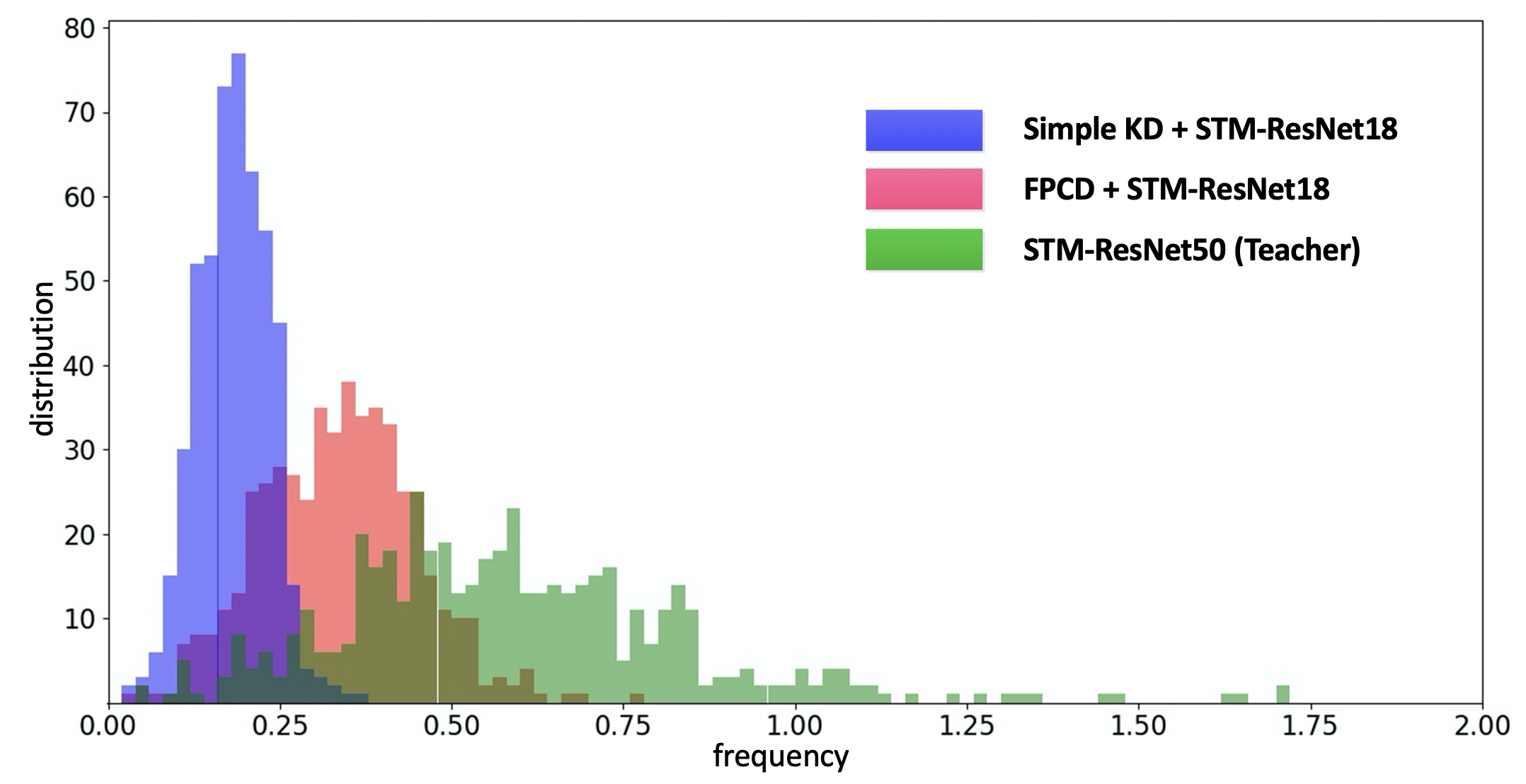}
	\caption{Illustration of the parameter frequency distribution of student networks with different knowledge distillation methods on STM \cite{jiang2019stm} framework. }
	\label{vis}
	   \vspace{-0.4cm}
\end{figure}

We continue to conduct the knowledge distillation performance comparisons of our FPCD with other different distillation methods \cite{hinton2015distilling,2019Correlation}, and the video classification results on several datasets including both the temporal-related datasets (i.e. Something-Something V1\&V2 \cite{goyal2017something} and Jester) and the scene-related datasets (i.e. Kinetics-400 \cite{carreira2017quo}) are shown in Table \ref{stm}. We can find that the two existing knowledge distillation methods reproduced by us for action recognition task effectively promote the performance of original STM-ResNet18. And our FPCD applied on STM-ResNet18 surpass these two previous distillation methods significantly and consistently among all public action recognition benchmarks. For example, FPCD gains 2.1\%, 3.0\%, 0.5\% and 2.3\% performance improvement of the top-1 accuracy on Something-Something V1, V2, Jester and Kinetics-400 datasets respectively compared with \cite{hinton2015distilling}, where only the Softmax output is considered for distillation. Besides, the performance of student network optimized with our FPCD method can even exceed the teacher model on Jester dataset.


\setlength{\tabcolsep}{1pt}
\begin{table}[t]
   \setlength{\abovecaptionskip}{-0.1cm} 
    \caption{Top-1 accuracy comparison between FPCD and other distillation methods (i.e. Simple KD \cite{hinton2015distilling} and CCKD \cite{2019Correlation}) on Something-Something, Jester and Kinetics-400 datasets respectively with TSN \cite{wang2016temporal}  framework .}
        \begin{center}
        \begin{tabular}{ccccccc}
        \hline
         Method &Backbone & SS-v1& SS-v2 & Jester & Kin-400 \\
        \hline\hline
        TSN & ResNet-50  &   19.7  &    27.8     &   81.0 & 66.8    \\
        \hline
        TSN & ResNet-18  &  15.0  &   26.3      &  79.5 & 61.4   \\
        Simple KD + TSN & ResNet-18  &    16.3       &       27.4       &  80.6 & 63.3   \\
              CCKD + TSN & ResNet-18  &    16.6      &      27.7      &  81.1 & 63.9   \\
        \textbf{FPCD} + TSN& ResNet-18 & \textbf{18.7} &\textbf{29.8} & \textbf{82.4} & \textbf{65.3} \\
          \hline       
    \end{tabular}
    \end{center}
    \label{tsn}
\end{table}


 \subsection{Ablation study}
As described above, the scene representation contributes a lot for determining the action category in most of the videos on the scene-related dataset, i.e. Kinetics-400. Hence, the spatial frequency information is more critical than the temporal frequency dynamics. We claim that the high-frequency distillation makes little sense on this kind of dataset which is responsible for capturing the short-term motion information, while the low-frequency distillation can pay much attention to the appearance clues. As shown in Table \ref{fsd}, we can explicitly find that the low-frequency distillation of the feature spectrum distillation strategy applied on STM framework achieves more performance gains on Kinetics-400 than the high-frequency distillation by 1.3\%, which demonstrates the rationality of our assumption. Moreover, we also visualize the low and high frequencies of the original RGB images respectively as shown in Fig. \ref{overview}, which illustrates that the low frequency describes scene relevant information, while the high frequency involves more distinct motion edges. 

To validate the effectiveness of each distillation term proposed in FPCD, we conduct a series of ablation experiments on several datasets and the results are shown in Table \ref{fpcd}. We can conclude that the Feature Spectrum Distillation (FSD) contributes more performance gains than the Parameter Distribution Distillation (PDD) on temporal-related dataset (i.e. Something-Something V1 \& V2), which demonstrates the importance of temporal frequency for this kind of datasets. However, the PDD module brings more significant improvement than the FSD module on the scene-related dataset (i.e. Kinetics-400), which attends to the spatial appearance clues most. Besides, the two distillation strategies together with the collaborative learning strategy can further increase the performance consistently. Furthermore, we compare the parameter frequency distribution of teacher (STM-ResNet50) and student (STM-ResNet18) networks with different knowledge distillation methods including FPCD and \cite{hinton2015distilling}, as shown in Fig. \ref{vis}. We can find that the student model distilled with our method has more similar parameter frequency distribution to the teacher model than \cite{hinton2015distilling}.


\setlength{\tabcolsep}{3pt}
\begin{table}[t]
\centering
  \setlength{\belowcaptionskip}{-0.2cm} 
  \setlength{\abovecaptionskip}{-0.1cm} 
    \caption{Top-1 accuracy comparison between FPCD and other distillation methods (i.e. Simple KD \cite{hinton2015distilling} and CCKD \cite{2019Correlation}) on Something-Something V1 and Kinetics-400 datasets respectively with I3D \cite{carreira2017quo} framework.}
        \begin{center}
        \begin{tabular}{cccc}
        \hline
         Method &Backbone & SS-V1& Kinetics-400 \\
        \hline\hline
        I3D & Inception V1  &   41.6  &    71.1       \\    
         \hline 
          I3D & Channel-halved  &    36.5    &    61.7   \\
        Simple KD + I3D & Channel-halved  &     37.4     &     63.4   \\
                CCKD + I3D & Channel-halved     &   38.0    & 64.1   \\
        \textbf{FPCD} + I3D & Channel-halved & \textbf{39.1} &\textbf{64.7}  \\
          \hline       
    \end{tabular}
    \end{center}
    \label{i3d}
   \vspace{-0.5cm}
\end{table}

 \vspace{-0.1cm}
\subsection{Generalizability evaluation}
To validate the generalizability of our FPCD for knowledge distillation, we continue to analyze the effect of different video frameworks. In addition to the STM, we conduct experiments with several typical video networks (i.e. TSN \cite{wang2016temporal} and I3D \cite{carreira2017quo}) using ResNet-18 as backbone on the four aforementioned datasets. And the evaluation results are shown in Table \ref{tsn} and Table \ref{i3d} respectively. In Table \ref{tsn}, we can find that the performance of our FPCD applied on TSN-ResNet18 can achieve competitive results on all temporal-related datasets and even surpass the teacher network with deep backbone  (ResNet-50) on Jester and Something-Something V2 datasets. And the performance on the scene-related dataset only exists 1.5\% gap between the teacher and student networks. Besides, on the lightweight Inception V1 backbone (channel-halved) of I3D framework, we can still obtain the consistent performance improvement on every dataset as shown in Table \ref{i3d}. To conclude, the extensive experiments reveal that our FPCD can achieve great generalizability to train a lightweight video network on different frameworks.

\vspace{-0.1cm}
\section{Conclusion}

We propose a novel video knowledge distillation framework (FPCD) for video action recognition, which includes two main distillation strategies. Firstly, the temporal frequency spectrums calculated based on video features are distilled into the student network for learning the motion and scene representation simultaneously with a spectrum loss. Besides, the frequency distribution of network parameters are further adopted as another signal for guiding the appearance modeling process, which is independent of the video datasets. Finally, the two distillation terms are combined with the standard classification loss for optimization through the proposed collaborative learning strategy, where redundant knowledge are eliminated adaptively during the distillation process. Extensive experiments are conducted on several public action recognition datasets with different video networks, which demonstrate the consistent effectiveness and efficiency of our proposed video distillation method.

\clearpage
%
%
\bibliographystyle{splncs04}
\bibliography{egbib}
\end{document}